\documentclass[]{spie}  %>>> use for US letter paper
\usepackage{amsmath,amsfonts,amssymb}
\usepackage{graphicx}
\usepackage[colorlinks=true, allcolors=blue]{hyperref}

\title{Improving Emergency Response during Hurricane Season using Computer Vision}

\author{Marc Bosch}
\author{Christian Conroy}
\author{Benjamin Ortiz}
\author{Philip Bogden}
\affil{Accenture, 800 N. Glebe Rd, Arlington, VA, USA}
\authorinfo{Further author information: (Send correspondence to Marc Bosch)\\Marc Bosch: E-mail: marc.bosch.ruiz@accenturefederal.com}

\begin{document} 
\maketitle

\begin{abstract}
We have developed a framework for crisis response and management that incorporates the latest technologies in computer vision (CV), inland flood prediction, damage assessment and data visualization. The framework uses data collected before, during, and after the crisis to enable rapid and informed decision making during all phases of disaster response. Our computer-vision model analyzes spaceborne and airborne imagery to detect relevant features during and after a natural disaster and creates metadata that is transformed into actionable information through web-accessible mapping tools. In particular, we have designed an ensemble of models to identify features including water, roads, buildings, and vegetation from the imagery. We have investigated techniques to bootstrap and reduce dependency on large data annotation efforts by adding use of open source labels including OpenStreetMaps and adding complementary data sources including Height Above Nearest Drainage (HAND) as a side channel to the network’s input to encourage it to learn other features orthogonal to visual characteristics. Modeling efforts include modification of connected U-Nets for (1) semantic segmentation, (2) flood line detection, and (3) 
for damage assessment. In particular for the case of damage assessment, we added a second encoder to U-Net so that it could learn pre-event and post-event image features simultaneously. Through this method, the network is able to learn the difference between the pre- and post-disaster images, and therefore more effectively classify the level of damage. We have validated our approaches using publicly available data from the National Oceanic and Atmospheric Administration (NOAA)’s Remote Sensing Division, which displays the city and street-level details as mosaic tile images as well as data released as part of the Xview2 challenge. In addition, we have integrated the CV-generated artifacts and results in a collection of analytic tools including routing, damage assessment, and response prioritization, to assist with response management and strategic decision. The routing tool allows users to plan optimal alternate routes given automatically detected flood lines from the latest imagery. Response prioritization estimators look for critical areas, e.g., homes surrounded by water, flooded roads, and other anomalies detected in the impacted area. Finally, the damage assessment tool includes a pixel-based financial model
capable of outputting estimated financial damage costs projected according to the United States National Grid (USNG) coordinate system. In conclusion, we are working towards an emergency response system that provides stakeholders timely access to comprehensive, relevant, and reliable information. The faster emergency personnel are able to analyze, disseminate, and act on key information, the more effective and timelier their response will be and the greater the benefit to affected populations.
\end{abstract}

\keywords{Computer vision, flood line detection, geospatial image analysis, natural disaster response.}

\section{INTRODUCTION}
\label{sec:intro}  % \label{} allows reference to this section
\begin{figure} [ht]
   \begin{center}
   \begin{tabular}{c} 
   \includegraphics[height=7.5cm]{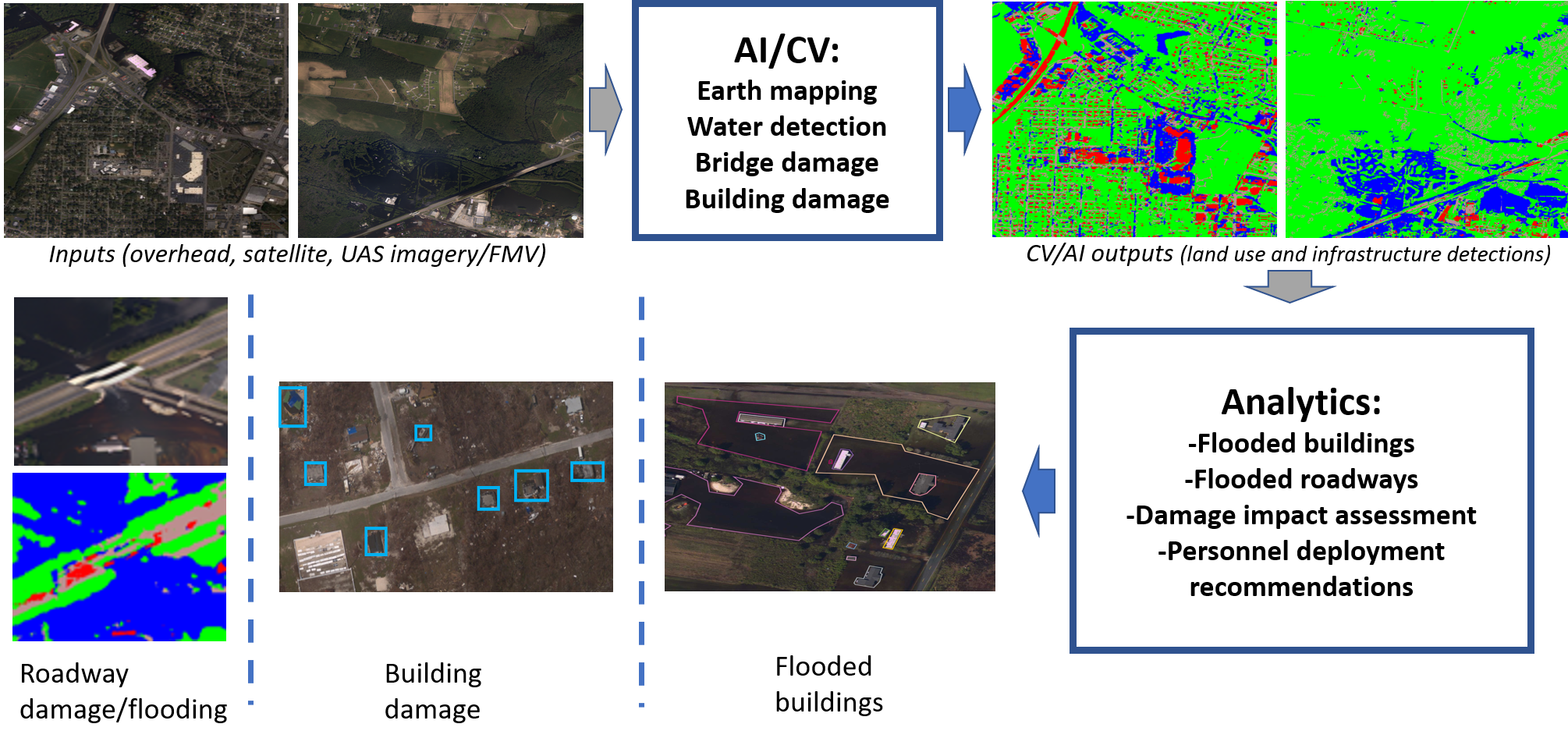}
	\end{tabular}
	\end{center}
   \caption[Example.] 
   { \label{fig:block} Block diagram for natural disaster relief.}
   \end{figure}

Floods are one of the most devastating and costly natural disasters, posing a significant threat to human life and property, and necessitating systematic and timely response to flood risks. While most floods cannot be prevented, they can be detected, and a quick response can greatly reduce the consequences.\\
According to the Centre for Research on the Epidemiology of Disasters (CRED), 280 floods occurred globally from 2008-2018, making flooding the most common natural disaster by far, and one of the most devastating~\cite{flood}. As a result of climate change and increasing global surface temperatures, the possibility of more floods, droughts, and increased intensity of storms has increased. More powerful storms are likely to develop as a result of an increase in the amount of water vapor being evaporated in the atmosphere. 
 Recently there has been several scholarly activities exploring the application of computer vision (along with wireless sensor data technology) to help equip flood management authorities with flood response and management solutions ~\cite{gebre,sarker,gaofeng,kahn}.\\
 In addition, federal agencies also conduct damage assessments to determine the extent of damage in a region following a natural disaster. These are often unsafe and laborious processes. Government agencies presently manual search and annotate these images for damage or obstructions. Automated processing of space-borne imagery offers the possibility to accelerate damage assessment.\\
 Thus, in this work we present a framework that leverages computer vision (CV) and analytics to estimate (1) semantics of pre/post-event imagery, (2) flood line detection, and (3) damage assessment using earth observation data.\\
 Our overall approach to humanitarian assistance and disaster relief using image analysis has two steps: (a) semantic mapping of the scene, (b) information extraction using analytics (e.g. flooded infrastructure, building damage, routing alternatives, personnel deployment recommendations, etc.). See Fig.~\ref{fig:block} for more details.\\
 Finally, we are interested in computationally efficient architectures. These algorithms must process large images in cloud environments using public funds, thus it is important to design solutions that include accelerators such as auxiliary data or include architectures that can solve multiple tasks (e.g. classification and localization) in one pass.
\section{U-Nets - Baseline Architecture for semantic analysis of overhead imagery}
\begin{figure} [ht]
   \begin{center}
   \begin{tabular}{c} 
   \includegraphics[height=12cm]{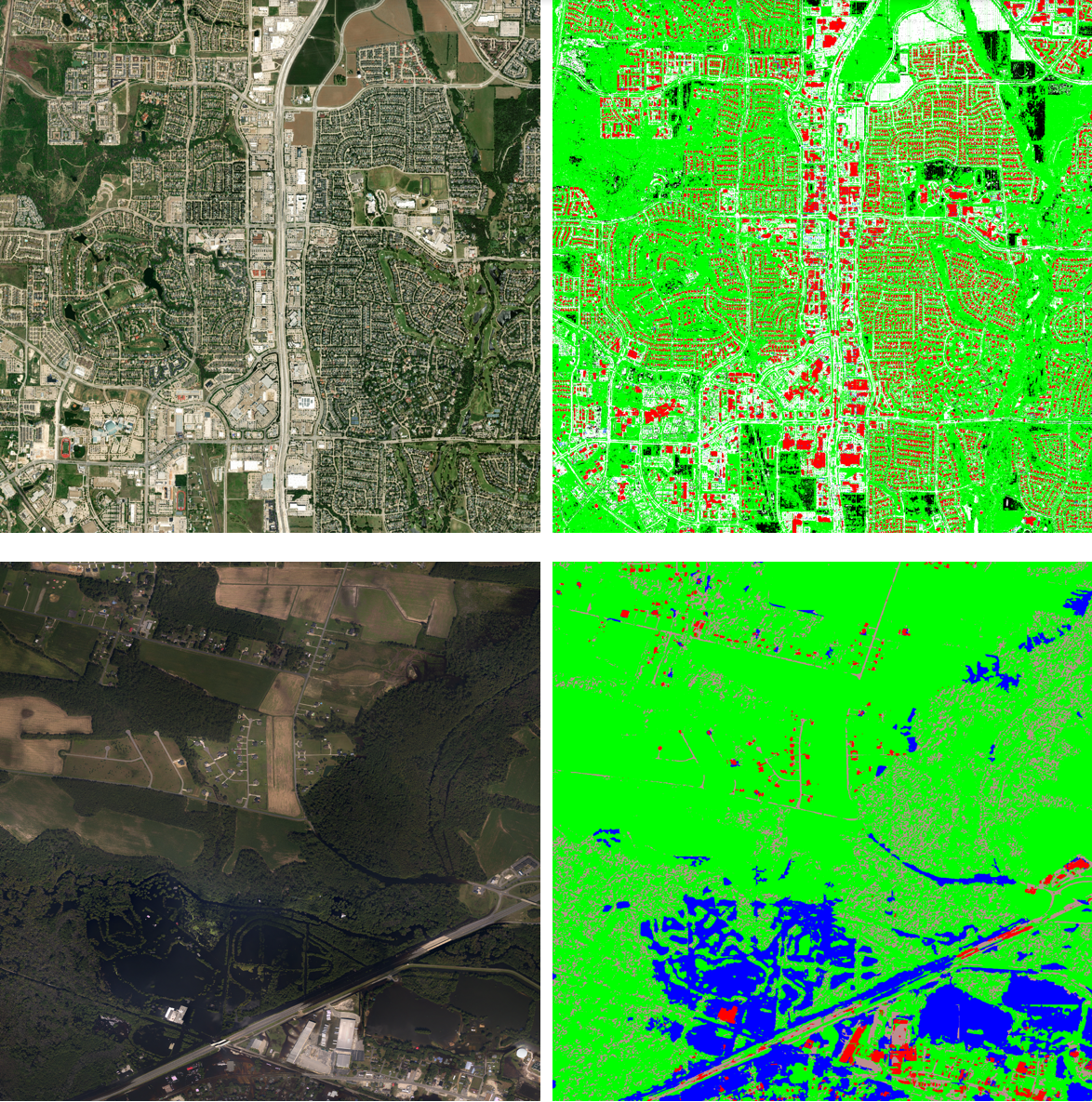}
	\end{tabular}
	\end{center}
   \caption[example] 
   { \label{fig:segment} Two examples of semantic segmentation for scene mapping using U-Nets.}
   \end{figure}

We selected the U-Net architecture as our base neural network ~\cite{unet} for this work. U-Net was originally proposed for biomedical segmentation, but it quickly gained attention in other domains of image analysis. Since in our domain label scarcity is a prevalent issue we needed an architecture that can leverage strong use of data augmentations. Ronneberger \textit{et. al.} proposed the U-Net architecture as a network that can use data more efficiently, while addressing shortcomings of predecessor architectures regarding localization. U-Net consists of a symmetric encoder-decoder structure, which enables precise localization and assignment of a class label to each pixel in the image. Unlike many other visual tasks, the U-Net’s architecture is more organized, works with very few images, produces more accurate segmentations, and can take less than a second to process a 512x512 image on a consumer grade GPU~\cite{unet}. Finally, U-Nets are a specific neural network architecture that can learn from the data to identify features of interest at the pixel level by leveraging multiple feature map resolutions.\\
%The next component of our tool applies semantic segmentation, a computer vision (CV) technique which adds near real-time overhead imagery information (\cite{McInness}, 2018). We use publicly available data from the National Oceanic and Atmospheric Administration (NOAA)’s Remote Sensing Division, which displays city and street-level images of Lumberton, NC as mosaic tile images. 
We used U-Nets to perform semantic segmentation to classify, at the pixel level, images acquired before and after a natural disaster. U-Nets map each pixel into categories of interest such as water, vegetation and man-made infrastructure \cite{Ishida}, \cite{Bosch}, \cite{Prasad}, \cite{Van Etten}. Figure \ref{fig:segment} shows two examples of semantic segmentation using U-Nets on two different locations where different semantic categories are mapped to the imagery at the pixel level.\\
\subsection{Adding Side Information to U-Net to Assist in Flood Line Detection}
\begin{figure}[h]
\centering\includegraphics[width=1\linewidth]{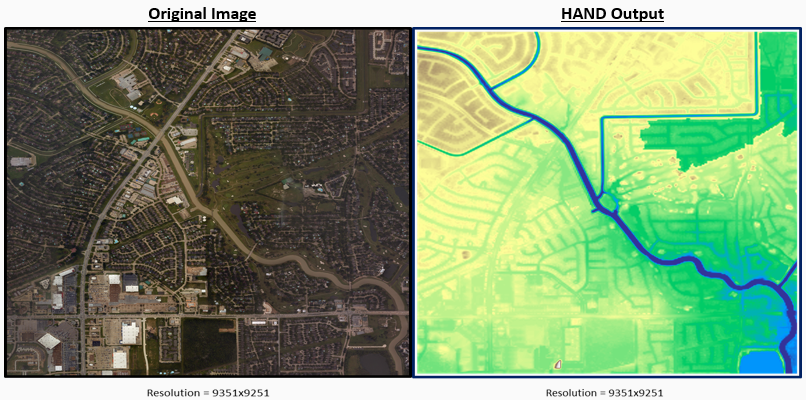}
\caption{Examples of aligned imagery and HAND data.}
\label{handdata}
\end{figure}

One of the bottlenecks in many computer vision frameworks which rely on supervised learning is the acquisition of annotated data that can be used as training data. One way to alleviate this need is to leverage other data sources that can lead to faster convergence during training or act as ``soft-guides'' to learn the task at hand with less data. In this particular case, the objective task is detection of flooded areas (after Hurricanes). We addressed flood line detection by using semantic segmentation in combination of data that encodes terrain information to facilitate detection with less training data and/or noisy training data. Height above nearest drainage point information is a relatively recent data layer that has been shown to be a good proxy for terrain elevation~\cite{hand1,hand2,hand3}.\\
Terrain-based approaches for inundation mapping have proven to produce accurate maps from a known streamflow and carefully created data layers. One of these data sources is the Height Above Nearest Drainage (HAND) technique~\cite{hand1}. HAND is a terrain model that normalizes input national elevation datasets (NED) according to the local relative elevations/heights found along a drainage network. It stores information regarding elevation data, discharge-height relationships, and streamflow inputs.\\
Once we obtain HAND data for a specific region (guided by the imagery geolocation), we proceed to combine HAND data and imagery, which involves augmenting the images by stacking an extra ``channel'' to the input tensor, which results in a four channel tensor with RGB channels along with the HAND output values contained within the HAND images. As for our algorithm and architecture, we used a U-Net architecture, as mentioned earlier, and experimented using different encoder backbones. The EfficientNet B-2 backbone~\cite{efficient} proved to be the best suited for our needs since it has a scaling method that uniformly scales all dimensions of depth/width/resolution using a simple yet highly effective compound coefficient, which offers favors tasks with different data source inputs. Moreover, we applied a series of pre-processing data augmentations and transformations including image rotations, brightness/contrast adjustments, and normalization applied per separate to RGB and HAND channels. Figure~\ref{handdata} shows an example of aligned HAND data with the source imagery.

\subsection{Two Encoder U-Net for Pre- and Post-event Analysis for Building Damage Assessment}
\label{sec:title}
\begin{figure}[h!]
\centering\includegraphics[width=0.65\linewidth]{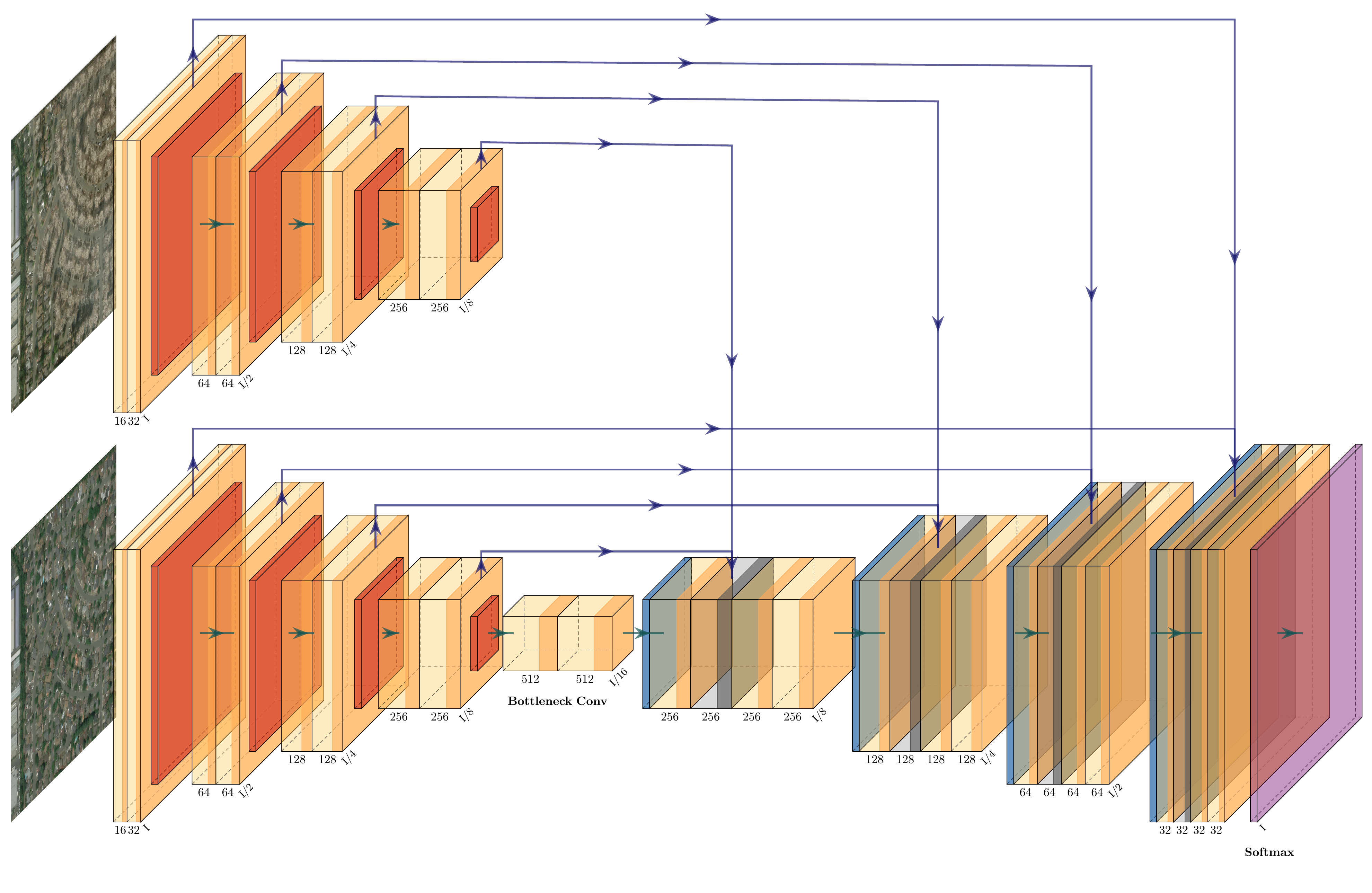}
\caption{U-Net architecture used for damage assessment.}
\label{fig:unet}
\end{figure}

Typically, conventional U-Net architectures, for the task of semantic segmentation, take an image as input and return that same image segmented into regions based on the classes specified during the network initialization. This strategy only enables to process one image at a time. This would force us to use one U-Net to analyze the pre-event image to locate buildings and another network to analyze the post-event image to predict the level of damage. In applications where change detection and characterization is part of the reasoning task, the network should be able to simultaneously analyze both images. To remedy this issue, we implemented a U-Net architecture with an additional encoder branch, allowing for dual image input. This not only streamlines the inference process, but enhances the training process as well. Through this method, the network is able to learn the difference between the pre- and post-disaster images. Both encoders share weights even though each image is convolved by different layers. At the decoder side, the layers are concatenated with the difference between the outputs of the pre- and post-disaster image encoder layers. The output of the model is a tensor of shape (1024, 1024, 5), with each channel in the final dimension representing one of the distinct damage levels: No Damage, Minor Damage, Major Damage, and Destroyed plus one class for a background. A visual representation of the network is shown in Figure~\ref{fig:unet}.\\%The entire network was implemented from scratch using TensorFlow and Keras.
During training, we encountered highly unbalanced training data (most buildings are not damaged in two different temporal snapshots). We addressed the class unbalance problem by combining the generalized Dice loss\cite{a9} with cross entropy with class weights. Using weights with cross entropy allowed us to place more of a penalty on the misclassification of certain classes to avoid the network to ignore under represented classes. Originally, predictive performance on minor and major damage was poor; therefore, any image containing pixels in either of these classes was oversampled to assist in learning.
\section{Results}
\subsection{Datasets}
We focused on two data sources. (1) Aerial imagery was obtained from NOAA NGS Coastal Image Services, and (2) satellite imagery from xBD dataset~\cite{xbd}. NOAA aerial images consisted of 4 bands and had a 9,351 x 9,351 resolution. We focused in data collected before and after hurricanes Harvey in Texas and Florence in North Carolina. We added the HAND channel to it for different areas of interest. For example, the HAND image covering the Harvey affected area had a 61,439 x 67,714 resolution. After projecting HAND data into the imagery, we tiled the source images to $1024 \times 1024$ pixels as input to the U-Net to match xBD inputs.
In contrast, the xBD dataset contains images across 19 natural disasters consisting of events such as volcanic eruptions, fires, floods, hurricanes, etc. Each image has also a resolution of $1024 \times 1024$ pixels. The dataset also introduces a Joint Damage Scale, which is an attempt to create consistency by having a unified damage scale across natural disasters \cite{xbd}. Each pixel in the images is labeled on a scale of 0-3, corresponding to the amount of damage. However, in an effort to create a unified model that simultaneously performs localization and classification, we have slightly adjusted these metrics to the following: 0 = No Building, 1 = No Damage, 2 = Minor Damage, 3 = Major Damage, 4 = Destroyed.\\
\subsection{Experiments}
\begin{figure}[h!]
\centering\includegraphics[width=0.65\linewidth]{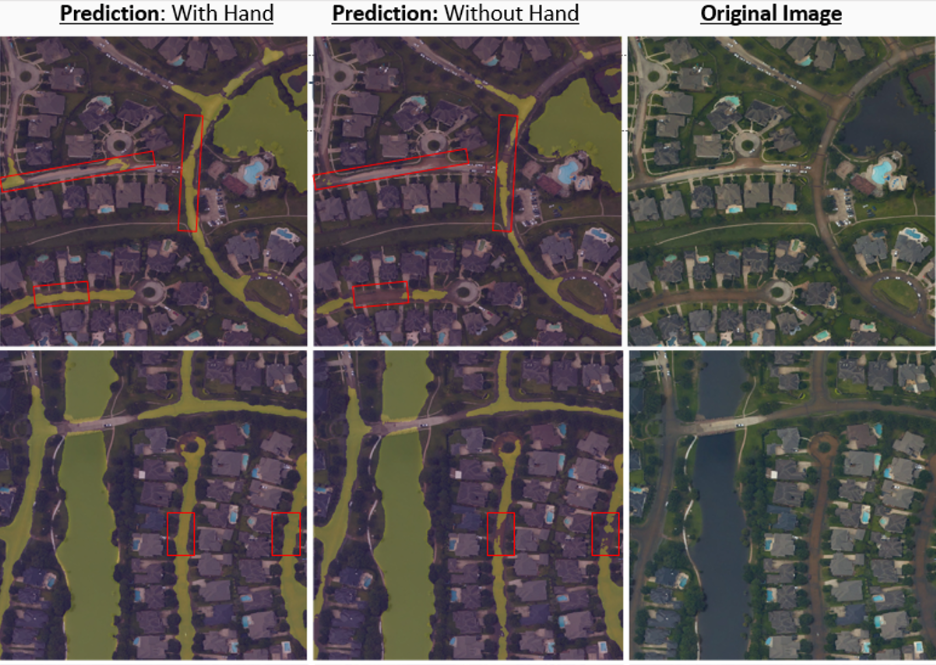}
\caption{Results of flooding detection with (left) and without (middle) HAND data.}
\label{handcomp}
\end{figure}

\begin{figure}[htb]
\centering\includegraphics[width=0.99\linewidth]{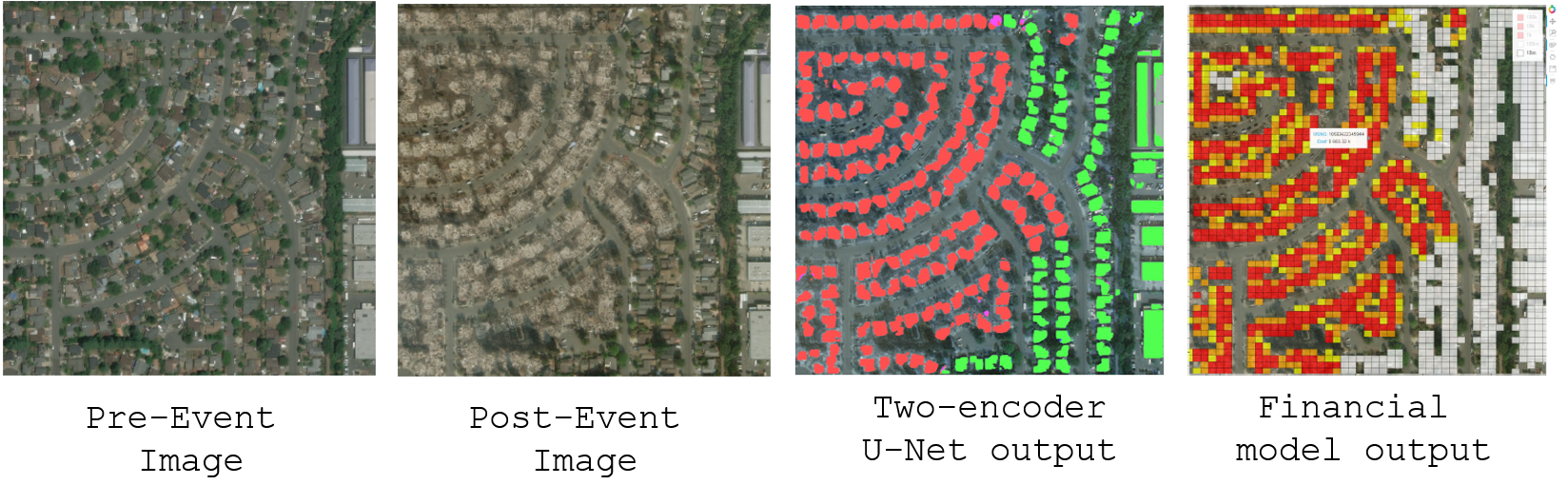}
\caption{Results of building damage assessment and financial model output.}
\label{bd}
\end{figure}

Comparing the performance of adding side information to a U-Net as a guide did not seem to bring consistent performance improvement in all scenes and environments. We believe that urban activity caused by temporal discrepancy between HAND data production and imagery acquisition introduced noise into the network that negatively impacted its performance. Nevertheless, we observed that adding HAND data seemed to help resolve cases of wet pavement vs. actual flooding and other areas where there was a noticeable improvement. Fig.~\ref{handcomp} shows different scene examples where RGB+HAND outperformed RGB-only cases for an area of Texas where Hurricane Harvey caused large damage. There is a task ahead of manually cleaning HAND data to filter out confusion. All in all, we believe that HAND is a good supplementary data source to improve flood line detection, expedite training and reduce the initial training set size.\\
Building damage assessment experiments were carried out using the xBD dataset. We compared our two-encoder U-Net architecture with the xBD baseline approach. Table~\ref{unetperf} shows the overall performance achieved with the proposed U-Net compared to the xDB baseline that included a two-step (building localization and damage classification) approach. Figure~\ref{bd} shows an example result of our damage assessment framework for damage level estimate given a pre- and post-event image.\\
\begin{table}[ht]
\begin{center}
\begin{tabular}{||c|c|c||}
\hline
\textbf{Metric} & \textbf{Our model} & \textbf{xDB Baseline} \\
\hline\hline
 F1 - Score & \textbf{0.616} & 0.265 \\ 
 F1 - Localization & \textbf{0.808} & N/A\\
 F1 - No Damage & \textbf{0.813} & 0.663\\
 F1 - Minor Damage & \textbf{0.365} & 0.144 \\
 F1 - Major Damage & \textbf{0.476} & 0.009\\
 F1 - Destroyed & \textbf{0.701} &  0.466\\ 
\hline
\end{tabular}
\end{center}
\caption{Performance Proposed Method For Building Damage Assessment.}
\label{unetperf}
\end{table}
\subsection{Large-Scale Visualization and Financial Damage Assessment}
\begin{figure}[htb]
\centering\includegraphics[width=0.65\linewidth]{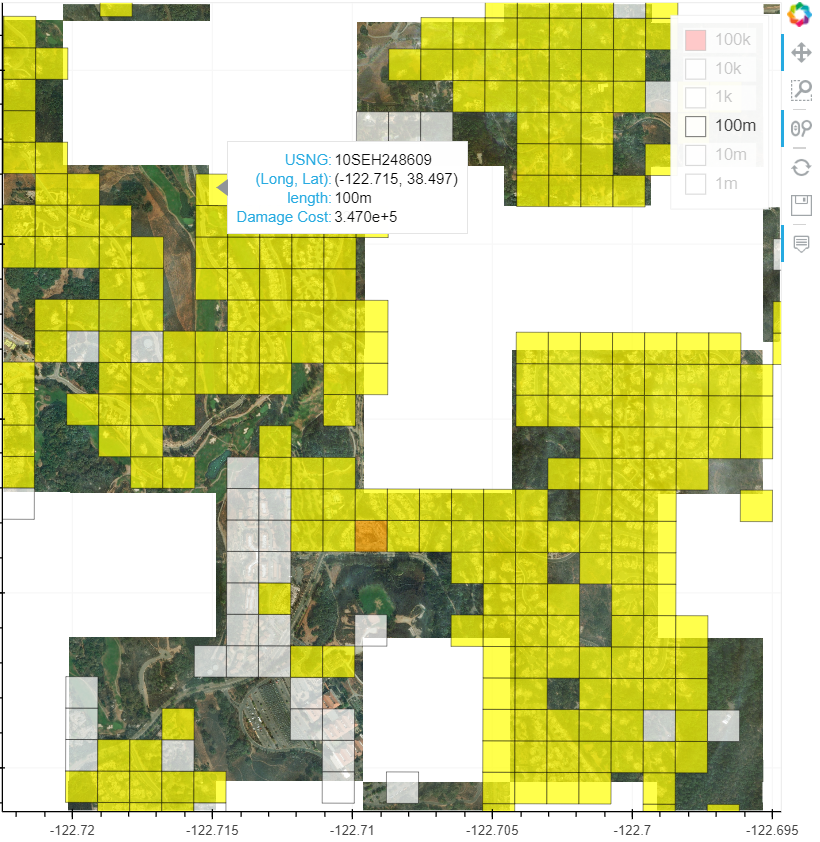}
\caption{Bokeh xBD U.S. National Grid Visualization}
\label{bokehplot}
\end{figure}

Once visual damage is estimated, we added a final step in order to calculate the expected cost associated with predicted damage levels. Figure~\ref{bd} (right) shows an example of financial impact based on the estimated damage level. A pixel-level financial model was developed based on building square footage. Real-state companies provide price per square footage costs for each zip-code. We used the 2018 Zillow Home Value Index (ZHVI) \cite{a14} for these set of experiments. Since imagery was georegistered this allowed to obtain the associated zip-code per detected building. The identified ZHVI cost per square-foot is multiplied  by the square footage of the footprint of the building and by a damage factor. The particular values of the damage factors were obtained by roughly validating the estimates against county property damage costs obtained from the National Oceanic and Atmospheric Administration (NOAA) \cite{a16}. The square footage of the building footprint is multiplied by two to account for multiple stories. The ability to rapidly project and damage estimates using different location systems is critical to summarize damage assessment results (\textit{e.g.} United States National Grid (USNG) coordinate system, county level, zip code, etc. ). In particular, the USNG is utilized by Government agencies to visualize and describe different information streams during and after disaster response. Integration of the output from computer vision (CV) automated assessment techniques into this coordinate system would speed up the analysis process and aid in providing fast, accurate, safely obtained, damage summaries. Using the technology capabilities developed during this effort, we have built a web application that uses satellite imagery to effectively estimate and summarize the fiscal damage to a region impacted by a natural disaster as shown in Fig.~\ref{bokehplot}.
\section{Conclusion}
Computer Vision and Image Analysis can become powerful accelerators for human assistance and disaster relief situations. In this work we focused on methodologies that provide general assistance and situational awareness (semantic mapping of the scene) during hurricanes, including flood line detection, and estimates of building damage up to a financial estimation. Our tools can be used in the context of managing risks in crisis and emergencies and can be used by different types of users to plan for natural hazards. We described several neural network architectures to solve the aforementioned tasks that can facilitate response during hurricane season. As well as mapping this layer of inferred knowledge to operational geospatial data representation frameworks like the U.S. National Grid for standardized description of information. We believe there is still much more that can be done with this technology but this is an initial step to help and improve human lives.
% References

\end{document}